# Unadapted Multilingual ASR on a Garrusi Kurdish Evaluation Set: A Common-Reference Staged Normalization Analysis

Hiwa Asadpour
Saarland University and Goethe University Frankfurt
asadpour@lingua.uni-frankfurt.de

## Abstract

Evaluating speech recognition for a Kurdish variety written in a Latin field orthography, using a model that outputs Arabic script, raises a measurement problem before it raises a modelling one. A direct comparison scores writing-system difference as recognition error. The usual remedy, normalizing reference and hypothesis together before scoring, changes the reference tokenization at the same time as the hypothesis, so the resulting improvement mixes an increase in agreement with a change in the denominator against which agreement is measured. I report an evaluation of MMS-1B-all with the Central Kurdish (ckb) adapter, used as released and without any adaptation on Garrusi data, on 1,722 segments of Phase 1 elicited questionnaire speech from five Garrusi speakers (9,763 reference word tokens, 117.9 minutes). Scoring uses a common-reference design: the reference is folded once and held fixed at 9,763 tokens across all conditions, and only the hypothesis representation varies. Scoring the unmodified Arabic-script hypothesis against the fixed Latin reference gives 111.70% WER and 100.92% CER with zero word-level matches. Transliterating the hypothesis to Latin gives 102.36% and 57.89%. Additionally folding it into the reference's reduced orthography gives 97.85% WER and 51.20% CER. Under this scoring design, the RAW-to-FOLDED transformation reduces the measured WER by 13.85 percentage points and the measured CER by 49.72 percentage points, of which the folding step contributes 4.51 and 6.69 respectively. Substantial measured error remains after both transformations: 14.53% of reference tokens are aligned as exact matches, the edits are substitution-dominated, and per-segment WER is higher for segments with fewer reference tokens. A Southern Kurdish fine-tuned system (aranemini/southern-kurdish-asr) scored under the same design performs worse on every speaker, with WER of 109.56% and CER of 55.85% over 1,703 segments, though 12,330 characters of its output fall outside the folding table and the run must be repeated against the corrected reference before its rates are final. Part of that residual is a scoring-pipeline artefact of unmeasured size rather than recognition error: the transliterator leaves 613 characters of the MMS hypothesis unconverted or unmapped, and 12,330 characters of the Southern Kurdish system's output, before folding (§6). I will release the fixed reference and the segment-level results, subject to the data-sharing terms of the source corpus, so that the scoring design can be checked independently. The Southern Kurdish system was scored against a pre-correction reference; its rates will be recomputed against the fixed reference before final publication.

## 1 Introduction

Kurdish speech technology has extended beyond its original concentration on Central Kurdish, with dedicated evaluation sets now available for Northern Kurdish, Badini, and Southern Kurdish (Mohammadamini & Tahon, 2026; Mohammadamini et al., 2026), and with recent work addressing variation within Central Kurdish itself (Ahmadi et al., 2024). For Southern Kurdish the most substantial of these is the resource of Mohammadamini and Tahon (2026), which provides 30 hours of validated read speech from 208 speakers together with a multi-domain benchmark of 100 sentences, translated from an existing Central Kurdish test set and read by eight speakers drawn from five vernaculars, and which releases the corpus, a fine-tuned MP32Vec2-BERT CTC model and inference code publicly. That study also reports a result of direct relevance to the configuration evaluated here: a Whisper model fine-tuned for Central Kurdish reaches 84.23 WER on their Southern Kurdish benchmark against 13.40 WER on the Central Kurdish version of the same sentences, which they read as evidence that systems developed for one Kurdish dialect transfer poorly to another. Coverage remains uneven, and the unevenness appears to follow the field's dialect labels: varieties not straightforwardly covered by the Sorani/Kurmanji labelling are correspondingly less visible in the published record (Asadpour, forthcoming). The dependence on labels is unsurprising given that the classification of Kurdish varieties is itself unsettled and is drawn on different criteria by different authors (Haig & Öpengin, 2014; Asadpour, 2021); §2 sets out the background. Garrusi is one variety that falls outside the two best-covered labels: it is described in recent linguistic work as a minority Kurdish variety of Iran (Asadpour & Zarei, 2026a, b), and it is not listed separately in the classification I cite here. Garrusi is not entirely absent from that record: the Southern Kurdish benchmark of Mohammadamini and Tahon (2026) includes one speaker labelled Garrusi among its eight, contributing 95 of its 773 recordings, alongside speakers of the Kermānshāhi, Kalhori, Malekshāhi and Kolyā'i vernaculars (their Table 2). That study reports character error rate per speaker and does not aggregate results by vernacular. Since four of the five vernaculars are represented by a single speaker, speaker and vernacular effects are not separable within that design, and the authors attribute the observed spread across speakers to speaker-related characteristics and not to variety. I found no evaluation reporting recognition rates specific to Garrusi, and no evaluation set assembled for the variety[1].

[1] Searched Google Scholar, ACL Anthology, HAL, arXiv, IEEE Xplore, and Scopus through 12 August 2026 for the variety name and its spelling variants (Garrusi, Gerrûsî, Garrousi, Bijari, Bîcarî) in combination with speech recognition, ASR, WER, CER, corpus, MMS, wav2vec and Whisper. I found linguistic work on the variety, and one benchmark that includes a Garrusi speaker among several vernaculars (Mohammadamini & Tahon, 2026), but no evaluation reporting recognition rates for the variety as such.

Two obstacles stand between that gap and a usable first measurement. No Garrusi-trained recognition system exists, so an initial evaluation must transfer a model adapted to a related variety. There is also a measurement problem. The model emits Central Kurdish Arabic script, while my reference transcriptions are in a Latin field orthography (§2), so a direct comparison scores a writing-system difference as recognition error. The standard remedy is to normalize before scoring, but normalization is usually applied to both sides at once, which means the reported gain mixes an actual increase in agreement with a change in the reference tokenization. Recent work on Kurdish ASR has observed that word error rate is inflated relative to character error rate by this class of standardization difference (Mohammadamini & Tahon, 2026; Mohammadamini et al., 2026). Separating the increase in agreement from the change in reference tokenization is necessary for interpreting a Kurdish WER figure, not a refinement of it.

The pilot reported here addresses both. I evaluate MMS-1B-all (Pratap et al., 2024) with the Central Kurdish adapter on elicited Garrusi questionnaire speech, an adapter chosen because prior linguistic analysis found substantial morphosyntactic overlap between Garrusi and Central Kurdish, despite clear phonological and phonetic differences, and score it under a common-reference staged normalization: the reference is folded once, fixed, and reused unchanged across three conditions that differ only in how far the hypothesis has been converted toward it. Because the reference tokenization is constant, the measured differences between conditions arise from the hypothesis-side transformation. The fixed reference and the segment-level results will be released alongside the paper, subject to the data-sharing terms of the source corpus (§8), so that the design can be checked rather than taken on trust. The result is a zero-shot ASR measurement on this Garrusi evaluation set, together with an explicit statement and full reporting of the scoring design, so that the effect of each hypothesis-side transformation on the reported rate is visible rather than absorbed into a single normalized figure. This is not a benchmark in the sense of a released dataset with a shared protocol.

## 2 Background: Kurdish Varieties and Orthographic Setting

Descriptive work commonly distinguishes five Kurdish varieties: Northern Kurdish (Kurmanji), Central Kurdish (Sorani), Southern Kurdish, Gorani, and Zazaki (Haig & Öpengin, 2014). Northern and Central Kurdish are the least disputed members of this grouping, and they are also the two that speech technology has concentrated on.
The boundaries between these groups are not settled. Published classifications differ from one another, combine geographic, historical, social and linguistic criteria in varying proportions, and often do not state which criterion is doing the work. There is no consensus in the literature on how Kurdish should be defined or subdivided (Haig & Öpengin, 2014; Asadpour, 2021). The classifications differ in how they place Garrusi: the general classification of Haig and Öpengin (2014) does not list Garrusi separately, whereas Fattah's classification, as reported by Belelli (2019), lists Bijāri, also known as Garrusi, among the Southern Kurdish subgroups (p. 78). Belelli also describes the Bijār area as a Southern Kurdish enclave in a predominantly Central Kurdish environment (p. 74) and reports lexical items shared with neighboring Central Kurdish varieties (p. 88). The model transferred here is adapted to Central Kurdish.

Central Kurdish is itself internally varied, with regional varieties including Mukri, Hewlêrî, Silêmanî, Germiyanî and Sineyî (Haig & Öpengin, 2014; Asadpour, 2021). For this experiment, the consequence is interpretive: a model adapter labelled "Central Kurdish" should be understood as adapted to some portion of that range and not to all of it. I make no claim about which Central Kurdish varieties are or are not present in the model's training data, which I cannot audit.

Garrusi is a Kurdish variety of Iran. The linguistic description I rely on for the speech evaluated here is that of Asadpour and Zarei (2026a, b), whose speakers were recorded in the Mehraban District of Hamadan Province, where the recordings analyzed here were also collected. The general classification cited above does not list Garrusi separately, and Southern Kurdish is there taken to cover varieties such as Kelhuri, Feyli and Kirmashani, with boundaries against neighboring varieties that are among the less settled in the literature (Haig & Öpengin, 2014). I have not attempted a survey of how Garrusi is placed across the classification literature, and I therefore use "Garrusi" as the variety label for the speech evaluated here without attempting to resolve its broader classification.

Kurdish is written in more than one script. Central Kurdish is commonly written in an Arabic-based script, but Latin-based conventions are also in use, and orthographic conventions can differ even where the same variety is being represented (Ahmadi, 2020, p. 73; Asadpour, forthcoming). The reference transcriptions used here are in a Latin field orthography with phonemic diacritics; the model evaluated here emits Central Kurdish Arabic script. The two sides of the comparison are in different writing systems before recognition accuracy enters into it at all.

An unmodified comparison would therefore measure a writing-system difference together with recognition error. I score the same hypotheses under progressively normalized representations, holding the reference fixed, so that the change in measured agreement produced by each transformation is visible separately from the error that remains. This shows how much of the measured rate moves under orthographic processing. It does not partition the remaining error into orthographic and acoustic components, and I do not read it as doing so (§6).

# 3 Data and Methods

## 3.1 Data

The material is Phase 1 of a Garrusi Kurdish field corpus: elicited questionnaire speech, recorded as MP3 files, with time-aligned reference transcriptions in a Latin orthography using phonemic diacritics.[2] The five speakers evaluated here (CZ, FI, MR, MY, SK) are those the processing run completed, of the thirty in Phase 1. The selection was neither random nor designed: the run processed speakers in a fixed order (CZ, MY, SK, FI, MR), which is not the order in which they appear in the corpus directory, and did not cover the remainder. No log of the run was kept, so the point at which it stopped is inferred from the output rather than recorded. I report the five without treating them as a sample of Garrusi speakers, and give per-speaker results without comparing them. No speaker metadata were available for this phase. No Phase 2 or Phase 3 material is used, and the evaluated set contains no free or conversational speech.

The processing script excluded segments shorter than 0.3 s and segments whose reference field was empty or a placeholder. Of 1,765 segments in the reference alignment files, 43 were excluded, all of them for falling below the 0.3 s threshold; no segment was excluded for an empty reference. After these exclusions, and after removing 247 duplicate rows produced by a resumed processing run, deduplicated on the speaker and segment identifier (1,969 rows reduced to 1,722), the evaluation set comprises 1,722 segments, 9,763 reference word tokens, and 7,073.0 s (117.9 minutes) of scored audio. That is a mean of 5.67 reference words per segment, a median of 5, and approximately 1.38 reference words per second. There is no train/development/test split, because no training or fine-tuning was performed; the entire set is an evaluation set.

## 3.2 Recognition

I used facebook/mms-1b-all (Pratap et al., 2024) with the Central Kurdish (ckb) adapter loaded and the tokenizer target language set accordingly. MMS uses small per-language adapter modules inserted into the pretrained backbone (Pratap et al., 2024); all other components were left as released. Audio was loaded at 16 kHz, mono, and clipped to the reference time alignments. Decoding was greedy CTC decoding, frame-wise argmax over the model's output distribution, with no beam search and no external language model. Pratap et al. (2024) report that the multilingual benchmark results for their multi-domain model were obtained using n-gram language models trained on Common Crawl at inference (their Table 5). My figures are therefore not comparable to those results and should not be read as the best obtainable from this model.

I performed no training, fine-tuning, or adaptation on Garrusi data of any kind; the model and adapter were used as released. I cannot audit the contents of the MMS-1B-all pretraining corpus, so "zero-shot" here describes my procedure rather than a verified absence of related material upstream.

The mismatch between the model's training material and my data is not only one of variety. MMS-1B-all is fine-tuned on a mixture of MMS-lab, FLEURS, CommonVoice, VoxPopuli and MLS, and MMS-lab is derived from New Testament recordings (Pratap et al., 2024). I do not know which of these corpora contribute to the Central Kurdish component. My data are field recordings of elicited questionnaire speech distributed as MP3, which I take to differ from these sources in speaking style and recording channel as well as in linguistic variety. The reported rates should therefore not be attributed to variety mismatch alone, and I do not attempt to separate the three.

A second system was subsequently run on the same audio: aranemini/southern-kurdish-asr, the MP32Vec2-BERT CTC model fine-tuned on Southern Kurdish released by Mohammadamini and Tahon (2026), used as released and without adaptation on Garrusi data. It is included because Garrusi is placed among the Southern Kurdish subgroups by at least one classification (§2) and is represented by one speaker in that study's benchmark, and because the checkpoint is publicly available. Like MMS-1B-all it emits Arabic script, and its hypotheses were passed through the same transliteration and folding path and scored against the same folded reference text, so its result is reported in the FOLDED condition. The normalization function distributed with the model's inference script was not applied, so the hypothesis received no treatment beyond the path described in §3.3. Transferring the transliteration step to this system has one consequence worth noting at the outset. The character ۊ (U+06CA) represents a vowel specific to Southern Kurdish and is not part of the Central Kurdish inventory the transliterator was written for, and it occurs in the raw hypothesis of 170 of the 1,703 scored segments. Inference failed on 19 of the 1,722 segments; §4.5 reports what was scored.

## 3.3 Text transformations

The two sides of the comparison start in different representations. The reference is in the Latin field orthography of the source corpus, which uses diacritics to mark phonemic distinctions. The hypothesis is in the Central Kurdish

[2] I am grateful to Masoumeh Zarei for her research assistance on the fieldwork and data collection on which this corpus is based. The recognition experiments, the scoring design, and the analysis reported here are my own, as are any errors that remain.

Arabic script emitted by the model. Two transformations bring them toward a common representation, and they are applied to different sides.

Transliteration converts Arabic script to Latin, and is applied to the hypothesis only, using KLPT 0.1.7 (Ahmadi, 2020). The transliterator was instantiated as Transliterate (“Sorani”, “Arabic”, target_script=“Latin”) and applied per segment via its transliterate method; other constructor arguments were left at their defaults. Its purpose is to remove the writing-system difference, so that the two sides are at least in the same alphabet. The reference requires no transliteration, as it is already Latin. KLPT’s Latin target is a romanization convention and is not the same convention as the field orthography of the reference. Transliteration therefore reduces the representational gap without closing it.

Folding maps a Latin string onto a reduced inventory. The input is NFC-normalized and lowercased. Each character is then replaced according to the mapping below, every remaining character outside [a–z and space] is converted to whitespace, and whitespace is collapsed.

â→a ê→e î→i ô→o û→u ü→u á→a é→e è→e
ļ→l ř→r r̂→r
ḧ→h ğ→h
ʕ ʔ ’ ʾ ʿ ‘ ‘ ‘ ` → (deleted)
ç→c ş→s š→s ž→j

The mapping is deliberately lossy: it collapses diacritic distinctions and deletes pharyngeal and glottal marks, so pairs that differ phonemically between reference and hypothesis may be counted as matches after folding. Any character not listed and not already in [a–z and space] becomes whitespace, which splits the token containing it; I verified the character inventory of both the reference and the transliterated hypothesis against this table. On the reference side the only characters outside the table are punctuation. On the hypothesis side 613 characters fall outside it: 502 U+FFFD replacement characters, 74 Arabic characters that KLPT left unconverted (39 U+0626, 35 U+06D5), 36 Latin diacritics that KLPT emits but the table does not list (22 *ẍ*, 14 *ë*), and one digit. Each of these becomes whitespace in the FOLDED condition and splits the hypothesis token containing it; §6 states what this means for the reported rates.

Folding brings two Latin conventions onto a shared reduced alphabet so that differences of diacritic and pharyngeal notation do not count as errors. It is lossy in both directions, and §5 sets out what it collapses.
Folding is applied to the reference once and the result is then held fixed. Across the three conditions only the hypothesis representation changes: RAW leaves it in Arabic script, TRANSLIT applies transliteration, and FOLDED applies transliteration and then folding. The pipeline is therefore not identical normalization on both sides, since the hypothesis undergoes a transliteration step that the reference does not. Describing the two sides as receiving identical treatment would misdescribe the design.

## 3.4 The common-reference scoring design

The reference is folded once and thereafter held constant. All conditions score the same 1,722 segments against the same 9,763-token folded reference; only the hypothesis representation changes:

**RAW:** the Arabic-script hypothesis, unmodified.
**TRANSLIT:** the hypothesis after transliteration to Latin.
**FOLDED:** the hypothesis after transliteration and folding.

Scoring used jiwer 3.0.3 for word and character error rate, with the library’s default transformations. WER is the word-level edit distance (substitutions plus deletions plus insertions) divided by the number of reference word tokens. CER is the character-level edit distance divided by the number of reference characters, with inter-word spaces counted as characters. Both are pooled at the corpus level, total edits divided by the total reference count, rather than averaged over segments.

Because the reference is fixed, S + D + H = 9,763 in every condition. A reader can confirm from this that the reference length is constant across conditions, a necessary condition for the design. Confirming that the reference tokens themselves are identical requires the released reference file, which I provide alongside the segment-level results (§8).

# 4 Results

## 4.1 Main result

Figure 1 shows the composition of the three conditions against the fixed reference; Table 1 gives the exact counts.

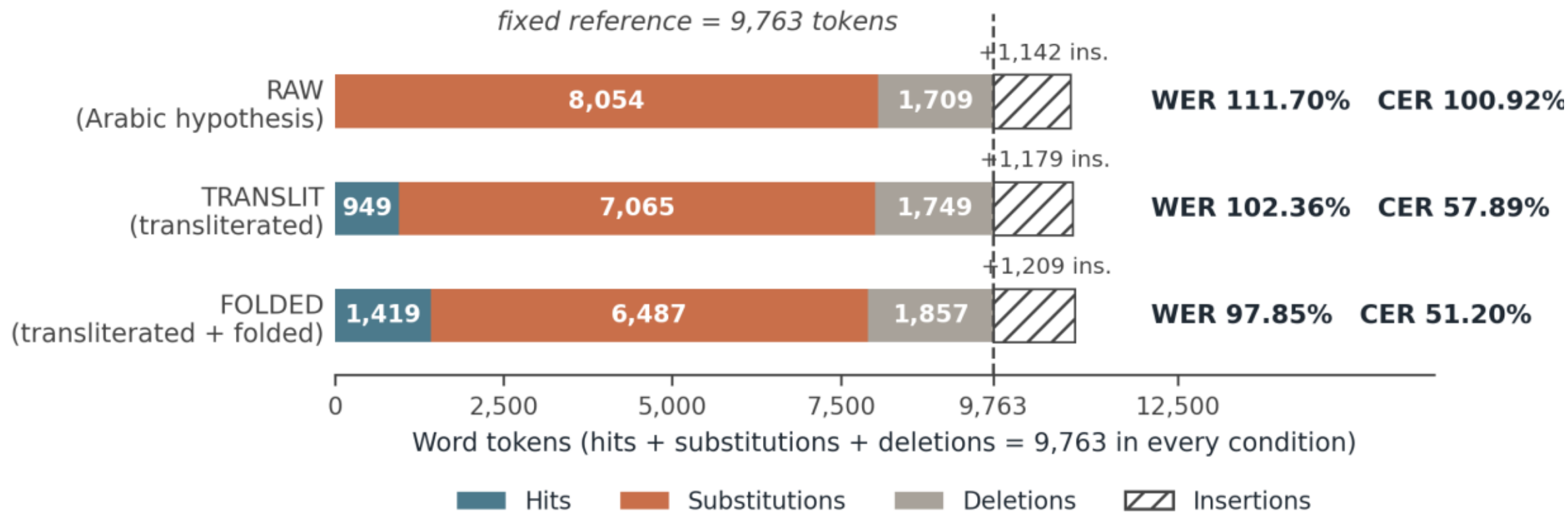

**Figure 1.** Alignment composition in each scoring condition, against the fixed 9,763-token reference. Hits, substitutions, and deletions sum to the reference length in every condition (dashed line); insertions extend beyond it, which is why WER exceeds 100% where hits are few. The figure shows that the two hypothesis-side transformations convert substitutions into hits while leaving deletions and insertions comparatively stable.

**Table 1.** Common-reference staged normalization. All three conditions score the same 1,722 segments against the same folded reference of 9,763 word tokens and 53,017 characters (inter-word spaces included, as counted by the scoring library); only the hypothesis representation differs. S = substitutions, D = deletions, I = insertions, H = hits. S + D + H = 9,763 at the word level in every row, confirming a constant reference length; identity of reference content across conditions is established by the released reference file, not by the table.

| Condition | Ref. tokens | WER | CER | S | D | I | H |
|---|---|---|---|---|---|---|---|
| RAW (Arabic hypothesis) | 9,763 | 111.70% | 100.92% | 8,054 | 1,709 | 1,142 | 0 |
| TRANSLIT (transliterated to Latin) | 9,763 | 102.36% | 57.89% | 7,065 | 1,749 | 1,179 | 949 |
| **FOLDED (transliterated + folded)** | **9,763** | **97.85%** | **51.20%** | **6,487** | **1,857** | **1,209** | **1,419** |

**Table 2.** Change in the measured rates between conditions.

| Transition | ΔWER | ΔCER |
|---|---|---|
| RAW → TRANSLIT | −9.34 pts | −43.03 pts |
| TRANSLIT → FOLDED | −4.51 pts | −6.69 pts |
| RAW → FOLDED | −13.85 pts | −49.72 pts |

The RAW condition returns zero word-level hits across all 1,722 segments. This follows from the comparison itself: a Latin-script reference and an Arabic-script hypothesis have disjoint token inventories, so no reference word can match. The resulting 111.70% is a scoring baseline for the ablation, not a recognition rate for the system. The final condition gives the main result: 97.85% WER and 51.20% CER. Of the 9,763 reference tokens, 1,419 (14.53%) are aligned as exact matches.

## 4.2 Error composition

In the FOLDED condition the 9,553 edits comprise 6,487 substitutions (67.9% of edits), 1,857 deletions (19.4%), and 1,209 insertions (12.7%). The hypothesis contains 9,115 word tokens, 93.4% of the reference length. The edits are substitution-dominated, and the system produces slightly less text than the reference instead of over-generating. Empty hypotheses were produced for 75 segments (4.4%). In the RAW condition, WER exceeds 100% because no reference token matches at all and insertions are added to a full complement of substitutions and deletions, not because insertions predominate.

## 4.3 Speaker-level results

**Table 3.** Speaker-level results, FOLDED condition. Figures recomputed per speaker from the segment-level data. All figures are computed against the corrected 9,763-token reference.

| Speaker | Segments | Ref. tokens | WER | CER |
|---|---|---|---|---|
| CZ | 230 | 1,261 | 101.11% | 53.92% |
| FI | 398 | 2,030 | 99.66% | 57.57% |
| MR | 307 | 2,176 | 90.62% | 43.74% |
| MY | 363 | 1,948 | 96.51% | 47.13% |
| SK | 424 | 2,348 | 102.34% | 54.43% |
| **All** | **1,722** | **9,763** | **97.85%** | **51.20%** |

Word error rate ranges from 90.62% to 102.34% and character error rate from 43.74% to 57.57%, so the pooled figure does not rest on a single outlying speaker. I do not interpret the differences between speakers. With five speakers, no metadata for this phase, and no analysis of which elicitation items each speaker contributed, the variation cannot be attributed to speaker characteristics, elicitation content, or recording conditions.

### 4.4 Segment length

Per-segment WER correlates negatively with the number of reference tokens in the segment (Spearman rho = −0.390, $p = 1.74 \times 10^{-63}$; Pearson r = −0.250, $p = 5.27 \times 10^{-26}$). Its association with segment duration is weak, and the two coefficients do not agree in sign (Spearman rho = −0.059, p = 0.015; Pearson r = 0.145, $p = 1.62 \times 10^{-9}$), so I do not treat duration as showing a consistent association in this set. Per-segment WER tends to be higher for segments with fewer reference tokens (Figure 2).

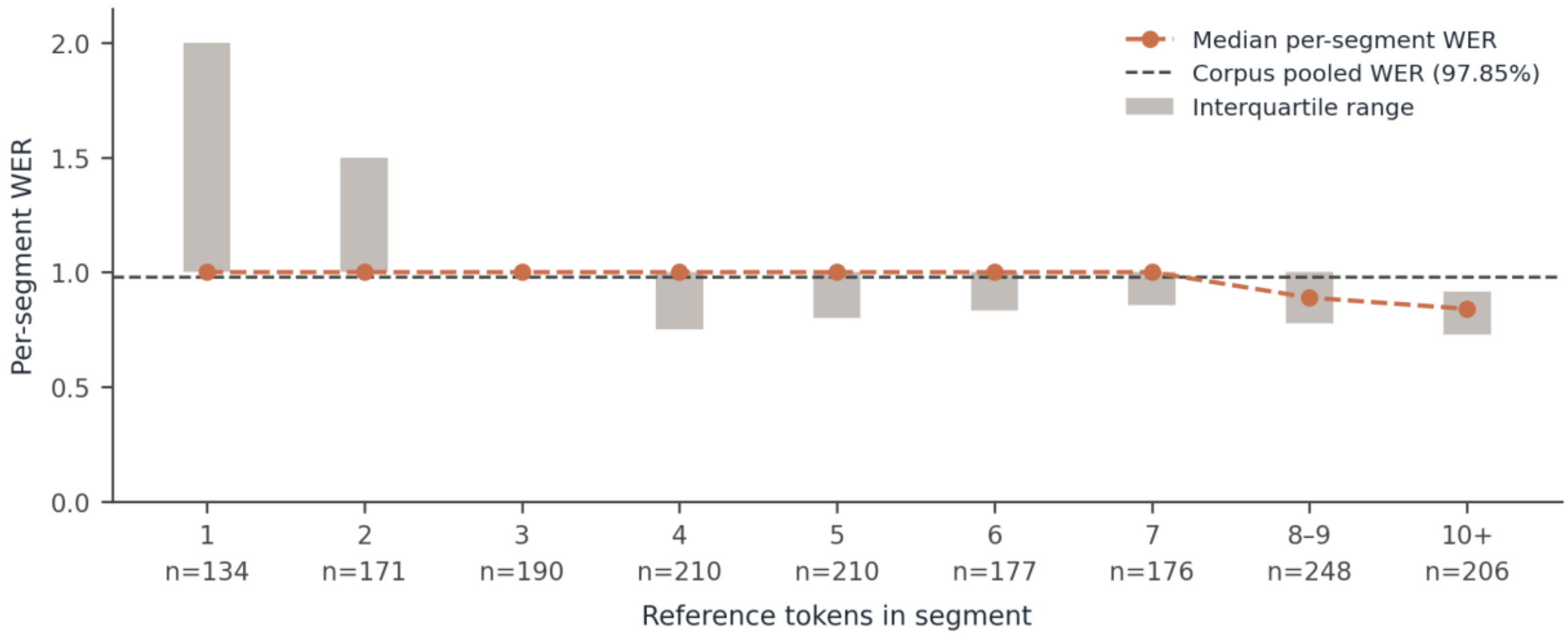

**Figure 2.** Per-segment WER by reference-segment length. Points are bin medians, bars the interquartile range (not the full range); n per bin is shown along the axis. Median WER is exactly 1.00 in every bin from one to seven reference tokens (n = 134, 171, 190, 210, 210, 177 and 176 respectively, together 1,268 segments or 73.6% of the set) and falls only for longer segments (median 0.889 for 8–9 tokens, n = 248; median 0.840 for 10 or more, n = 206). The qualifications in the text apply: part of this pattern follows from the definition of the metric rather than from recognition difficulty.

Three qualifications apply: First, because per-segment WER is defined with the reference token count as its denominator, part of this association is a property of the metric and not of recognition difficulty. In very short segments the rate takes few, widely spaced values and exceeds 1.0 readily, whereas longer segments yield values concentrated nearer the corpus rate. I therefore report the relationship descriptively and do not treat it as an estimate of a length effect on recognition. Second, the token-count and duration coefficients are not independent measurements: reference token count and duration are themselves strongly correlated in this set (Spearman rho = 0.654, $p = 5.71 \times 10^{-211}$). The contrast between them indicates which of two correlated length measures tracks per-segment WER more closely, not that duration is unrelated to recognition difficulty. Third, with n = 1,722 the p-value for the duration coefficient reflects sample size rather than the magnitude of the association.

The set is composed of short segments: median duration 3.0 s, a median of 5 reference tokens, and 495 of the 1,722 segments (28.7%) containing three or fewer reference tokens.

### 4.5 A Southern Kurdish fine-tuned system

The Southern Kurdish fine-tuned system described in §3.2 was run on the same audio and scored in the FOLDED condition against the same corrected reference text, though not over the same segments. Inference failed on 19 segments (FI 13, MR 2, MY 2, SK 2), leaving 1,703 of the 1,722 scored. These are inference failures rather than exclusions under the criteria of §3.1, and no empty hypothesis was substituted for them. The two systems are scored over segment sets differing by 19 segments and over correspondingly different reference token counts, so the rows of Table 4 are not a within-design comparison in the sense of Table 1. I report the difference between them as a direction and not as a measured gap.

**Table 4.** The Southern Kurdish fine-tuned system alongside the FOLDED MMS-1B condition. The rows are scored over segment sets differing by 19 segments; see the qualifications in this section and in §6.

| System | Segments | Ref. tokens | WER | CER | S | D | I | H |
|---|---|---|---|---|---|---|---|---|
| MMS-1B-all, ckb adapter (FOLDED) | 1,722 | 9,763 | 97.85% | 51.20% | 6,487 | 1,857 | 1,209 | 1,419 |
| aranemini/southern-kurdish-asr | 1,703 | 9,912 | 109.56% | 55.85% | 6,539 | 709 | 3,612 | 2,664 |

The measured rates are higher for the Southern Kurdish system on every speaker and on both metrics. Against the per-speaker figures of Table 3, word error rate is 123.29% for CZ against 101.11%, 112.29% for FI against 99.66%, 100.32% for MR against 90.62%, 107.34% for MY against 96.51%, and 110.29% for SK against 102.34%; character error rate is likewise higher for each. The two systems order the speakers similarly at the extremes, with MR lowest and MY second lowest under both.

The error composition differs more than the pooled rate does. Of the Southern Kurdish system's 10,860 edits, 6,539 are substitutions (60.2%), 709 deletions (6.5%), and 3,612 insertions (33.3%); its hypothesis contains 12,815 word tokens, 129.3% of its reference length, and 2,664 reference tokens (26.9%) are aligned as exact matches. The corresponding figures for the FOLDED MMS condition are 67.9%, 19.4% and 12.7% of 9,553 edits, a hypothesis 93.4% of reference length, and 14.53% of reference tokens matched. The Southern Kurdish system therefore matches a substantially larger share of reference tokens while producing considerably more text than the reference, and the MMS adapter matches fewer while producing slightly less. Word error rate above 100% arises here for a different reason than in the RAW condition of §4.1: there it followed from the absence of any word-level agreement, whereas here insertions are added to a substantial complement of matches.

# 5 Discussion

**The measured rate and the 97.85% WER is a transfer result:** a model adapted to Central Kurdish, applied without any Garrusi data, on elicited speech recorded in the field and distributed as MP3. It characterizes one off-the-shelf configuration on this evaluation set, not the range of systems that might be applied to this variety. It does not indicate what Garrusi ASR could achieve with in-variety data. The closest published work in experimental shape is the Southern Kurdish benchmarking of Mohammadamini and Tahon (2026), which compares models tuned on Southern Kurdish against a Central Kurdish-tuned baseline. My configuration is likewise a Central Kurdish adapter evaluated on a variety outside the Central Kurdish standard, but I do not restate their figures here and do not attempt a numerical comparison with them, as the corpora, genres, and recording conditions differ. What is comparable is their released system, not their reported rate. Running their Southern Kurdish fine-tuned model on this evaluation set (§4.5) holds the audio, the segmentation and the transcriptions constant and varies only the recognizer. Their released checkpoint supports that comparison where their published figures do not.

There is a further reason for not attempting a numerical comparison, and it bears directly on the design adopted here. Mohammadamini and Tahon (2026) observe that the absence of a settled Southern Kurdish orthography allows the same word to be written in several forms, that this spreads small character-level differences across many word tokens, and that it therefore inflates WER relative to CER. Their error typology assigns a separate category to such standardization mismatches, in which both the reference form and the recognized form are attested spellings, and their reported figures are consistent with this account at 24.26 WER against 4.09 CER.

Their analysis is, in effect, independent evidence that a substantial share of a Kurdish word error rate is a property of the representation in which agreement is measured, not of recognition. What the published description does not state is the representation in which their own rates were computed: no normalization procedure, tokenization rule, treatment of punctuation and Unicode format controls, or scoring implementation is specified. This does not affect the validity of the figures within their stated evaluation setting, where the comparisons they draw involve differences of fifty WER points or more, but it does place two specific limits on what can be done with those figures. The first concerns reproduction: a reader who downloads the released corpus, checkpoint and inference script can transcribe the benchmark audio with the same model, but cannot establish that the processing applied to hypothesis and reference before alignment was the processing that produced 24.26 and 4.09, so the published rate cannot be recomputed from the released artefacts and checked. The second concerns the authors' own analysis. Having identified standardization variation as a source of the WER–CER divergence, and having isolated it as an error category in which both the reference form and the recognized form are attested spellings, the study establishes that this variation contributes to the reported WER without establishing how much of it would remain under a normalization that neutralized the category. That residual quantity is not recoverable from a single reported rate, and it is the quantity that a staged design measures directly. Holding the reference fixed and reporting each hypothesis-side transformation separately, as in §3.4, is one way of making that share visible rather than absorbing it into a single figure.

The structure of that benchmark separately determines which of its comparisons its evidence supports. It comprises 100 sentences read by eight speakers, of which 773 recordings passed validation, so its content dimension is one hundred sentences and the same text is scored eight times over. For the contrasts drawn between systems fine-tuned on Southern Kurdish and systems trained on other varieties, which involve differences of fifty word error rate points or more, this is not a material constraint: no plausible treatment of the repetition would reverse them, and the finding that models developed for other Kurdish varieties perform poorly on this material is securely established. Two further comparisons are more tightly bounded. The first is the ranking of the two fine-tuned systems, where MP32Vec-BERT reaches 24.26 WER and Whisper-Turbo 34.43 from single runs reported without repeated seeds or intervals. The authors offer two candidate explanations, the larger and more multilingual pretraining of the MP32Vec-BERT encoder and its character-level CTC prediction units, and do not separate them. The gap is therefore evidence that this configuration performs better on this benchmark, not evidence about which property produces the difference. The second is variety: four of the five vernaculars are represented by a single speaker, so the per-speaker character error rates, which range from 3.02 to 6.49, cannot be attributed to vernacular rather than to speaker, and the authors do not attribute them. The benchmark therefore establishes a cross-dialect transfer gap on Southern Kurdish read speech, and does not yet establish how recognition accuracy varies across the vernaculars it samples.

**What the staged normalization measures:** The experiment measures how word- and character-level agreement changes as the same hypotheses are progressively converted from Arabic script to a Latin representation and then folded into the reference's reduced orthography, with the reference tokenization held fixed throughout. Four observations follow directly. First, the RAW condition has no word-level agreement at all, so cross-script comparison without conversion cannot be interpreted as a recognition measurement. Second, the transliteration step recovers character-level agreement while leaving most word forms unmatched: character error rate falls from 100.92% to 57.89%, and exact word matches rise from none to 949 of the 9,763 reference tokens, with word error rate still above 100% at 102.36%. Third, folding reduced the measured rates by a further 4.51 WER and 6.69 CER points. Fourth, after both transformations, the measured WER is still 97.85%, and 85.47% of reference tokens are not matched.

I do not describe any part of this as isolating the causal share of error attributable to script mismatch. The transformations change the representation in which agreement is measured; they do not partition the underlying causes, and I do not claim that any specific number of WER points of the model's error originates in script mismatch. Folding in particular is lossy: it collapses diacritic distinctions and removes pharyngeals, so some pairs that differ phonemically between reference and hypothesis are counted as matches after folding, and some that differ only in an unmapped character are split into separate tokens. The FOLDED rates should therefore be read as performance under the stated reduced orthography, alongside the TRANSLIT condition rather than in place of it. What can be said is narrower: under this scoring design, the RAW-to-FOLDED transformation reduces the measured WER by 13.85 percentage points, and a large measured error remains once it has been applied.

**The residual:** The residual is large and substitution-dominated: substitutions account for 67.9% of the edits, hypothesis length is 93.4% of reference length, and 14.53% of reference tokens are aligned as exact matches. I do not read it as recognition failure alone: it may also contain orthographic and word-boundary differences that transliteration and folding do not resolve, and the present design does not identify what the remaining errors consist of. Whether some other orthographic treatment would reduce it further is untested.

A single segment illustrates one error type, a word-boundary difference counted as substitutions. It was selected to show that pattern and is not a typical segment: at 44.44% WER it is well below the corpus rate. In segment MR.141 (10.0 s, WER 44.44%, CER 19.61%), the folded reference

ew masin esbabbazige ki we tenabo kisirili ewe kewe<br>
is recognized as<br>
ew masin espebazige u tenaw bo kisirili ewe kewe

Both strings contain nine tokens; five align as exact matches and the remaining edits are four substitutions, with no deletions or insertions: several content words survive unchanged while the relative marker and the following prepositional phrase are realized as different word forms. Part of the difference is a word-boundary convention rather than a misrecognized word: the material the reference writes as the single token *tenabo* appears in the hypothesis as *tenaw bo*. Word-level scoring has no way to register this as a single boundary difference and counts it among the substitutions. How much of the corpus-level substitution count arises this way is not measured here.

**The second system:** A system fine-tuned on Southern Kurdish does not perform better on this Garrusi material than a Central Kurdish adapter used off the shelf; under the rates in §4.5 it performs worse, on all five speakers and on both metrics. The direction is consistent, and the two systems fail in different ways: the adapter produces slightly less text than the reference and deletes, while the fine-tuned system produces considerably more and inserts, matching a larger share of reference tokens while scoring the worse rate. Since the normalization function distributed with that model was not applied and no word-segmentation step was introduced on the hypothesis side (§3.2), the insertion excess is a property of what the system emitted rather than of how it was tokenized for scoring.

Three things restrict what this supports. First, the two runs are scored over segment sets differing by 19 segments and do not share a reference token count, so the size of the difference is not established, only its direction. Second, the systems are matched on nothing but the audio: they differ in architecture, training data, decoding, and pretraining, so a difference between them is not attributable to variety adaptation as such. Third, both hypotheses pass through a transliteration path specified against Central Kurdish Arabic script, while Southern Kurdish orthography includes characters that script does not use (§3.2); an unmeasured share of the second system's measured error may therefore be introduced by the scoring pipeline rather than by the recognizer, as §6 already argues for the first.

The two training materials also differ in ways this comparison does not separate from variety. The Southern Kurdish corpus is read speech: sentences of three to fifteen tokens drawn from five native writers and fifteen volumes of a magazine, revised by four professional editors who excluded non-Southern-Kurdish material and rewrote code-switching and loanwords in Kurdish script, recorded remotely through a messaging bot and a web tool, and validated for intelligibility, noise dominance and reading miscues (Mohammadamini & Tahon, 2026).

The material evaluated here is elicited questionnaire speech recorded in the field. A system trained on edited written sentences read aloud is therefore being asked to transcribe a different speaking style and a different recording channel, as well as a variety its own benchmark samples with a single speaker. I do not attempt to separate these contributions, and the result should not be read as a measurement of dialect distance.

What follows for evaluation is narrow but not trivial. A resource assembled from edited read sentences establishes performance on edited read sentences; whether a system trained on such material transfers to field-recorded speech of the same broad variety group is a separate question, and one that a benchmark drawn from the same collection protocol cannot answer. On this material it did not transfer well. That is a single observation on one evaluation set under one scoring design and it indicates where a measurement is missing without establishing what would fill it.

**Segment length:** That per-segment WER is higher for segments with fewer reference tokens is reported descriptively, with the qualifications in §4.4. Two explanations are available and neither is tested here. One is arithmetic: in short segments the rate moves in large steps and a single error produces a disproportionate value, in a segment of three reference tokens a single error costs 33 percentage points, and 28.7% of this set contains three or fewer reference tokens. The other is that shorter segments may offer less surrounding context to the recognizer; this is a possible explanation, not an observed result. Practically, the relationship suggests that segmentation policy is a variable worth holding constant in comparative Kurdish ASR evaluation, since two systems evaluated over differently segmented versions of the same audio will not be straightforwardly comparable.

**Exploratory analysis:** I do not interpret it further, and it is not among the paper's findings.

# 6 Limitations

**Scope of the evaluation set:** Five speakers were processed for the analysis reported here, of thirty in Phase 1, because a run that processed speakers in a fixed order did not continue past the fifth; no log was kept, so the interruption is inferred from the output rather than recorded. A later run of the same configuration over all thirty speakers, covering 8,240 of 8,650 segments and 47,160 reference tokens after excluding segments returning an empty hypothesis, gave a pooled word error rate of 102.15% and a character error rate of 57.94%. That figure is not directly comparable with the 97.85% reported here, since the present set retains its 75 empty hypotheses (§4.2) whereas the larger run excludes its 410. The difference this makes is small, of the order of a quarter of a percentage point on the five-speaker set, but the two rates are computed under different treatments. I do not report the larger run in the results because the staged conditions, the fixed reference, and the segment-level release were all constructed over the five-speaker set. The larger run indicates that the five speakers are not unrepresentative of Phase 1 in the aggregate. They are neither a random nor a designed sample, no speaker metadata were available for this phase, and no demographic or speaker-level interpretation is offered. The material is elicited questionnaire speech only: no Phase 2 or Phase 3 material, and no free, conversational, or naturalistic speech, is included, so the results do not extend to spontaneous Garrusi. There is no train/development/test split, since no training or fine-tuning was performed. The figures characterize this evaluation set under this configuration; they are not a benchmark result in the sense of a shared, released protocol.

**A correction to the folding table:** Folding reduces every character outside the reduced Latin inventory to whitespace, so any character absent from the mapping table splits a reference token rather than being normalized. An earlier version of the mapping omitted ü (U+00FC), which occurs 179 times in the reference across 165 segments and was therefore converted to whitespace. Of those occurrences, 170 are word-internal and each split one reference token into two; the remaining 9 are word-initial and split nothing. The reference count was thereby inflated from 9,763 to 9,933 tokens. The results reported here use the corrected mapping given in §3.3, in which *ü* is mapped to *u*. Folding remains lossy in the other direction, as described in §5: it collapses diacritic distinctions and removes pharyngeals, so the reference tokenization is not identical to the transcriber's even after correction.

**Transliteration was incomplete:** KLPT 0.1.7 did not fully convert the hypothesis to Latin: 502 characters of its output are U+FFFD replacement characters and a further 74 are Arabic characters left unconverted. The TRANSLIT condition is therefore not wholly in Latin script, and in the FOLDED condition these characters become whitespace and split the hypothesis tokens containing them. A further 36 Latin diacritics that KLPT emits (*ẍ, ë*) are absent from the folding table and are treated the same way. I did not correct this; the rates in Table 1 are computed with it present. The transliterator is also known to be imperfect on Kurdish independently of this: it is rule-based, and its detection of the unwritten vowel *i* was evaluated at 39% accuracy (Ahmadi, 2020, p. 77). Both defects belong to the scoring pipeline and not to the recognizer, and both mean the reported rates may overstate recognition error by an amount this study does not measure. I have not corrected the folding table or the transliteration fallback here, since doing so would change every reported figure; a future version of this evaluation should map these characters rather than discard them.

**The second system was scored over a smaller segment set, and through a transliterator not written for its output:** Inference with the Southern Kurdish fine-tuned model failed on 19 of the 1,722 segments, so the rates in §4.5 are computed over 1,703 segments and a correspondingly smaller reference than the 9,763-token reference used in Tables 1 to 3. The failures are not distributed evenly across speakers, falling mainly on FI, and I did not establish their cause; they are not exclusions under the criteria of §3.1 and no empty hypothesis was substituted for them. The consequence is that the two systems, though scored against the same reference text, are not scored against a reference of the same length, which is the property the design in §3.4 is intended to guarantee, and the comparison in §4.5 is accordingly weaker than the comparisons within Table 1 in a specific and remediable way. I report the direction, which is consistent across all five speakers and both metrics, and not the size of the difference. The transliteration limitation described above also applies with greater force to this system, since the Southern Kurdish orthography includes at least one vowel character absent from Central Kurdish (§3.2). 12,330 characters of its transliterated output fall outside the folding table and become whitespace, against 613 for the first system. Re-running the second system over the full set against the fixed reference is the correction required; neither point affects the figures for the first system, which are computed as reported throughout.

The staged normalization supports no causal claim, and no ceiling was measured. The design varies the hypothesis representation and holds the reference fixed. It establishes how much the measured agreement changes under each transformation, but not what causes the remaining errors, and it does not partition the reported rate into orthographic and acoustic components. In particular, I did not measure the agreement this pipeline would return for a hypothesis that was already correct. KLPT's Latin output and the reference field orthography are separate conventions, and folding narrows but does not necessarily close the gap between them, so an unknown share of the residual may originate in the scoring pipeline and not in the recognizer. Establishing this would require a control I did not run: passing a correct Arabic-script rendering of a sample of the reference sentences through the identical transliteration and folding path and scoring it against the fixed reference. I regard this control as a requirement for any stronger interpretation of the reported rates.

The exploratory annotation comparison (§5) should not be cited as a result: it was not significant at conventional levels, it used sentence-level morphological rather than syntactic labels, and it was computed over an incomplete join.

**Configuration specificity:** The source recordings used in this analysis are MP3 files; encoding parameters are not documented in the corpus metadata available to me. The segmentation follows the existing reference time alignments, and the system is a Central Kurdish adapter applied to a variety outside Central Kurdish. All of these are properties of this setup and not of Garrusi speech recognition in general.

**Comparability across Kurdish evaluations:** One limitation is not specific to this setup but constrains how any of these figures may be read alongside others. Reported Kurdish ASR rates currently differ in at least three respects that are not consistently documented. The first is segmentation: per-segment rates take the reference token count as their denominator, so two systems evaluated over differently segmented versions of the same audio are not straightforwardly comparable (§4.4). The second is the composition of the evaluation set, which determines how many independent samples a recording count actually represents (§5), and of which the five speakers reported here are an equally clear instance, being neither a random nor a designed sample. The third is the scoring representation, also discussed in §5. Each is a reasonable design choice under the constraints of a low-resource setting, but they are not documented to the same degree, and the difference has a practical consequence. The composition of the benchmark of Mohammadamini and Tahon (2026) is fully described in their Table 2, so a reader can judge for itself that its evidence supports conclusions about overall performance on Southern Kurdish read speech while leaving open performance on spontaneous or conversational speech, on broader speaker populations, and on each vernacular considered separately. The representation in which their rates were computed is not stated, so the corresponding judgement cannot be made at all: the reported rates remain interpretable within their evaluation setting and are not usable as a reference point outside it. My own figures have the reverse property, being computed under a representation that is stated in full but is idiosyncratic to this study, so they are reproducible and not comparable. In neither case is the underlying work at fault. In both cases a single WER figure for a Kurdish variety is a quantity that can be defended within a study and not carried between studies. The remedy is a reporting convention: state the segmentation, the composition of the evaluation set, and the representation in which scoring was performed, and release the fixed reference alongside the rate.

## 7 Ethics and consent

The recordings analyzed here were collected during the author's fieldwork on Garrusi Kurdish in collaboration with Masoumeh Zarei. All material was collected with the informed consent of the speakers and is stored in accordance with the General Data Protection Regulation on institutional servers. No new recordings were made for this study, and no participant was recontacted. No speaker metadata are reported in this paper, and no speaker is identified beyond an anonymized two-letter code.

## 8 Author contributions

The research concept, research question, technical design, implementation, analysis, interpretation, and writing of this study were developed and carried out by the author. The Garrusi fieldwork was conducted by Masoumeh Zarei under the author's supervision and with the author's assistance. Zarei carried out the initial transcription and translation of the recordings and provided updates during this process; the transcription and translation were subsequently checked, reviewed, and revised by the author as part of the project's quality-control process. The data processing, quality control, and preparation of the material for the present study were conducted by the author. No analysis or interpretation previously published by Zarei in the two joint publications was used in the present study. Zarei's contribution to the collection and initial preparation of the Garrusi data is acknowledged in this paper.

## 9 Data and code availability

The fixed 9,763-token folded reference and the segment-level results, per segment: speaker, segment identifier, timing, reference and hypothesis strings in each representation, and per-segment WER and CER, will be released alongside the paper, subject to the data-sharing restrictions attached to the source corpus. Releasing the fixed reference is what allows the common-reference design to be independently checked, since the S + D + H identity in Table 1 establishes only that reference length is constant across conditions. The released reference is the newline-joined concatenation of the 1,722 folded segments, with SHA256 340d66a0563363e0c48bb68ae158f7189f6eb730b57e016fa2efa33f772d0d3a, so that the exact file scored here can be identified.

The scoring scripts are released for a specific reason. Running a system and reproducing a published measurement are distinct properties of a release, and the Southern Kurdish resources of Mohammadamini and Tahon (2026) illustrate both sides of that distinction. Their corpus and their fine-tuned MP32Vec2-BERT CTC checkpoint are openly available on the Hugging Face Hub without registration or an access request, the corpus as eleven Parquet shards totaling 4.41 GB and the model as a 0.6-billion-parameter checkpoint distributed with its tokenizer, feature-extractor configuration and an inference script[3]. An independent researcher can therefore transcribe the benchmark audio with the same model, which is more than most published Kurdish ASR work currently permits. The published evaluation figures are a separate matter: recomputing 24.26 WER and 4.09 CER would additionally require the preprocessing applied to references and hypotheses before alignment and the implementation used to compute the rates, and neither is specified in the paper or distributed with the release. That release therefore supports independent inference but not independent reproduction of the reported metric, and it is the second property that determines whether a published rate can serve as a reference point against which later work is measured.

Software versions used: Python 3.11.9, transformers 4.35.0, torch 2.7.0+cu118, librosa 0.11.0, KLPT 0.1.7, jiwer 3.0.3, NumPy 2.2.6, SciPy 1.15.1. Model checkpoints: facebook/mms-1b-all, revision 3d33597edbdaaba14a8e858e2c8caa76e3cec0cd; aranemini/southern-kurdish-asr, revision 6debc819b2b3d482d23f089a6aa38f84f5f3b42d.

## 10 Conclusion

I report an ASR measurement for Garrusi Kurdish: MMS-1B-all with the Central Kurdish adapter, used as released and without adaptation on Garrusi data, yields 97.85% WER and 51.20% CER on 1,722 segments of elicited questionnaire speech from five speakers. Scoring under a common-reference design, in which the reference tokenization is fixed across conditions and only the hypothesis representation varies, the RAW-to-FOLDED transformation reduces the measured WER by 13.85 percentage points and the measured CER by 49.72 percentage points, with the folding step beyond transliteration contributing 4.51 and 6.69 respectively. A large measured error remains after both transformations. I do not attribute that residual wholly to recognition: without a measurement of what this pipeline returns for an already-correct hypothesis (§6), its composition is not established. I suggest that cross-script Kurdish ASR evaluation report the substitution, deletion, insertion, and hit counts and release the fixed reference alongside them, so that a reader can see how much of a reported rate moves under orthographic processing, how much does not, and against exactly which reference the rate was computed.

[3] Corpus: https://huggingface.co/datasets/aranemini/southern-kurdish-asr. Model: https://huggingface.co/aranemini/southern-kurdish-asr. The benchmark is additionally redistributed as the sdh subset of https://huggingface.co/datasets/aranemini/kurdish-multidialect-asr-benchmark. All three verified as publicly accessible without authentication on 16 August 2026; the model repository recorded 1,104 downloads in the preceding month.

Two further steps remain: One is the control described in §6, passing a correct Arabic-script rendering of a sample of the reference sentences through the identical transliteration and folding path, so that the agreement this pipeline returns for an already-correct hypothesis is measured rather than assumed. The second concerns the Southern Kurdish fine-tuned model of Mohammadamini and Tahon (2026), which I ran on this evaluation set with the result reported in §4.5: it did not perform better than the Central Kurdish adapter on this material, on any of the five speakers, and it failed in a different way, over-generating where the adapter deletes. That run covered 1,703 of the 1,722 segments and was scored against a pre-correction reference of 9,912 tokens, not the fixed 9,763-token reference (§6), so it establishes a direction and not a magnitude; repeating it over the full set under the same fixed reference remains outstanding. Because both systems emit Arabic script while the reference here is a Latin field orthography, such a comparison would require the scoring design set out above, and would allow the contribution of variety mismatch to be separated from that of the writing-system difference more directly than the present configuration permits.

A further limitation of that comparison is now measured. The transliteration step, which the Southern Kurdish system shares with the first, is more damaging for Southern Kurdish than for Central Kurdish: 12,330 characters of its output fall outside the folding table and become whitespace, against 613 for the MMS adapter, a twenty-fold difference (§4.5). Part of the Southern Kurdish system's apparent underperformance is therefore a scoring-pipeline artefact, not recognition error, and the two cannot be separated without the control described above.

One practical obstacle to the vernacular half of that comparison is worth recording, because it is straightforward to remove. Contrasting Garrusi against the other Southern Kurdish vernaculars on the existing benchmark requires grouping its recordings by variety, and the vernacular labels are reported in Table 2 of Mohammadamini and Tahon (2026) as a description of the eight speakers rather than carried as a grouping variable in the evaluation data as distributed. A per-vernacular rate is therefore a reconstruction, obtained by matching a printed table to the speaker identifiers in the release, not a computation over the released fields. he information exists and the reconstruction is not difficult. What matters is that the analysis the paper describes and the analysis the download supports are not the same, and carrying speaker-level vernacular labels in the distributed metadata would make per-vernacular evaluation, including for Garrusi, directly available to anyone using the benchmark.

The measurement problem this pilot isolates is therefore not only a technical matter of scoring design but also a data-infrastructure matter: without a fixed, released reference and without vernacular labels carried in the evaluation metadata, the same model can be reported with different rates against different references, and per-variety comparison remains a reconstruction rather than a computation.